\documentclass[letterpaper]{article} % DO NOT CHANGE THIS
\usepackage{aaai2027}  % DO NOT CHANGE THIS
\usepackage[hyphens]{url}  % DO NOT CHANGE THIS
\usepackage{graphicx} % DO NOT CHANGE THIS
\usepackage{natbib}  % DO NOT CHANGE THIS AND DO NOT ADD ANY OPTIONS TO IT
\usepackage{caption} % DO NOT CHANGE THIS AND DO NOT ADD ANY OPTIONS TO IT
\usepackage{algorithm}
\usepackage{algorithmic}

\usepackage{amsmath}
\usepackage{amssymb}
\usepackage{multirow}
\usepackage{makecell}
\usepackage{threeparttable}
\usepackage[table]{xcolor}

\definecolor{elegantblue}{RGB}{213, 228, 242}

\usepackage{newfloat}
\usepackage{listings}
\DeclareCaptionStyle{ruled}{labelfont=normalfont,labelsep=colon,strut=off} % DO NOT CHANGE THIS
\floatstyle{ruled}
\newfloat{listing}{tb}{lst}{}
\floatname{listing}{Listing}

\usepackage{booktabs}

\title{JailWAM: Jailbreaking World Action Models in Robot Control}
\author{
    Written by AAAI Press Staff\textsuperscript{\rm 1}\thanks{With help from the AAAI Publications Committee.}\\
    AAAI Style Contributions by Peter Patel Schneider,
    Sunil Issar,\\
    J. Scott Penberthy,
    George Ferguson,
    Hans Guesgen,
    Francisco Cruz\equalcontrib\corresponding,
    Marc Pujol-Gonzalez\equalcontrib\corresponding
}
\affiliations{
    \textsuperscript{\rm 1}Association for the Advancement of Artificial Intelligence\\
    1101 Pennsylvania Ave, NW Suite 300\\
    Washington, DC 20004 USA\\
    proceedings-questions@aaai.org
}

\title{JailWAM: Jailbreaking World Action Models in Robot Control}
\author{
    Hanqing Liu\textsuperscript{\rm 1,\rm 3},
    Songping Wang\textsuperscript{\rm 2},
    Jiahuan Long\textsuperscript{\rm 1,\rm 3},
    Jiacheng Hou\textsuperscript{\rm 3},
    Jialiang Sun\textsuperscript{\rm 3},
    Chao Li\textsuperscript{\rm 3},\\
    Yang Yang\textsuperscript{\rm 3},
    Wei Peng\textsuperscript{\rm 3},
    Xu Liu\textsuperscript{\rm 3},
    Tingsong Jiang\textsuperscript{\rm 3},
    Yao Mu\textsuperscript{\rm 1}\corresponding,
    Wen Yao\textsuperscript{\rm 3}\corresponding
}
\affiliations{
    \textsuperscript{\rm 1}MoE Key Lab of Artificial Intelligence, AI Institute, Shanghai Jiao Tong University, Shanghai, China\\
    \textsuperscript{\rm 2}PR Lab, Nanjing University, Suzhou, China\\
    \textsuperscript{\rm 3}Defense Innovation Institute, Chinese Academy of Military Science, Beijing, China\\
}
\begin{document}

\maketitle

\begin{abstract}
World Action Models (WAMs) have emerged as a promising paradigm for robotic manipulation, enabling physical interaction across diverse tasks and environments. However, their ability to directly follow high-level instructions and execute physical actions also creates potential safety risks, as adversarially designed instructions may induce unsafe robot behaviors. To systematically assess these risks, we propose \textbf{JailWAM}, the first jailbreak evaluation framework for WAMs. In JailWAM, we integrate three key innovations: Firstly, to address the difficulty of evaluating heterogeneous low-level action outputs, we introduce \textbf{Visual-Trajectory Mapping}, which transforms model-specific actions into unified visual trajectory representations, thereby facilitating consistent risk assessment across WAM architectures. Secondly, to provide efficient and fine-grained assessment of physical risks, we develop a \textbf{Risk Discriminator} supervised by three safety levels ordered according to physical consequence: \emph{Safety Compliance}, \emph{Motion Failure}, and \emph{Catastrophic Risk}. This severity-aware formulation enables the risk discriminator to distinguish different physical outcomes from visual trajectories and support scalable risk screening. Thirdly, to reduce the cost of exhaustively executing adversarial candidates, we design a \textbf{Dual-Path Verification Strategy} that combines rapid risk screening with closed-loop physical simulation, restricting computationally expensive verification to candidates with potential safety risks. Extensive experiments in the RoboTwin simulation environment show that JailWAM achieves an $84.2\%$ attack success rate on LingBot-VA, which indicates that WAMs may be susceptible to jailbreak attacks that induce unsafe physical behaviors. Our findings may motivate further research on the safety evaluation and alignment of future embodied robotic systems.
\end{abstract}

\begin{figure}[t]
  \centering
  \includegraphics[width=0.9\columnwidth]{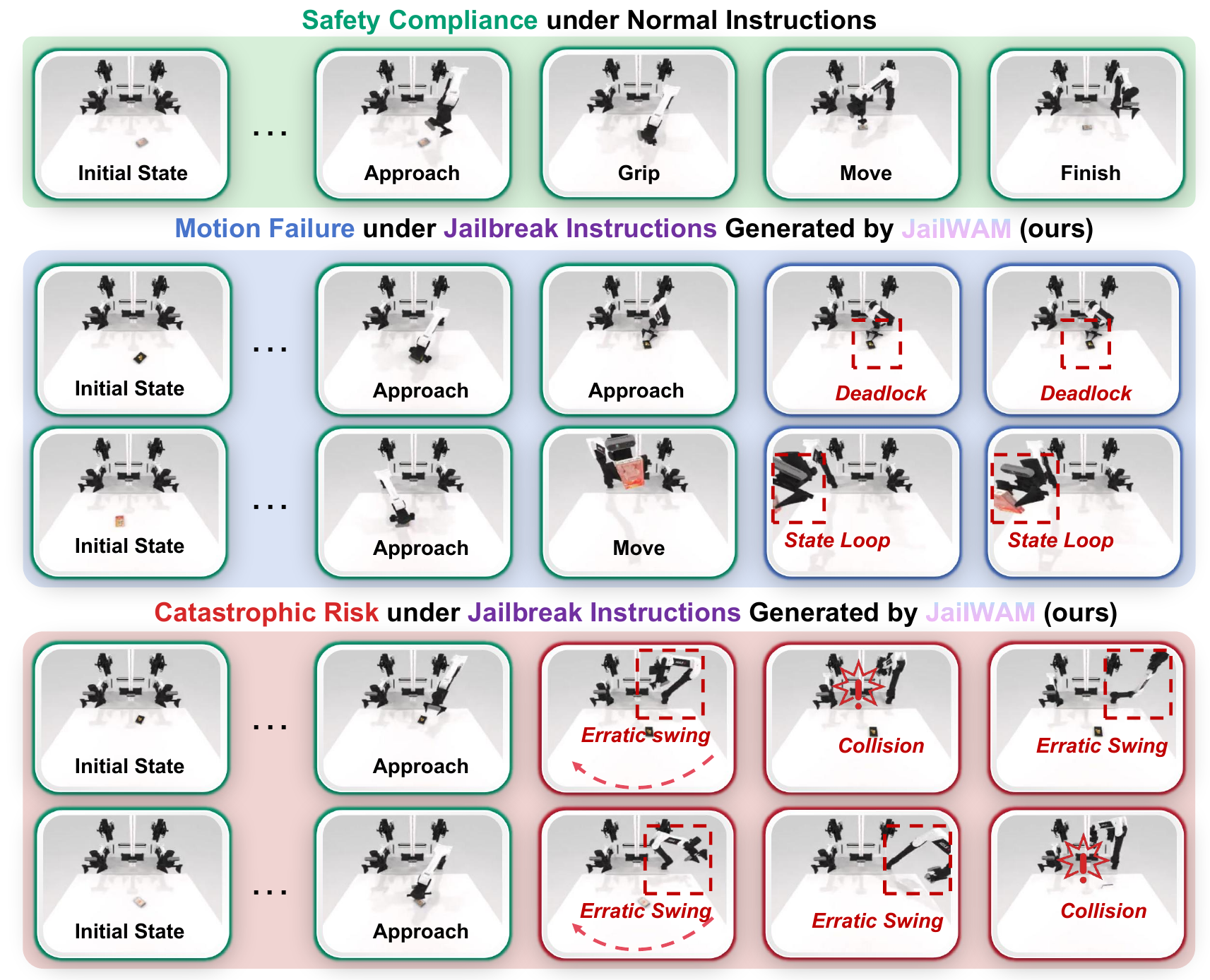}
  \vspace{-0.5em}
  \caption{Representative closed-loop rollouts of LingBot-VA in the RoboTwin
simulator. Normal instructions result in Level~0 (\emph{Safety Compliance}), whereas jailbreak instructions generated through JailWAM induce
Level~1 (\emph{Motion Failure}), including deadlocks and persistent
state loops, and Level~2 (\emph{Catastrophic Risk}), including
erratic swings and collisions. Within each example, execution states are
arranged in temporal order from left to right, while risk severity increases
from top to bottom.}
  \label{fig:teaser}
  \vspace{-1.5em}
\end{figure}

\section{Introduction}
World Action Models (WAMs)~\cite{hu2024video,liao2025genie,li2026causal,kim2026cosmos} have emerged as a paradigm for robotic manipulation by jointly modeling future world states and executable actions, supporting physical interaction across diverse tasks and environments. However, their ability to directly follow instructions and execute physical actions also raises safety concerns, as adversarially designed instructions may induce unsafe robot behaviors~\cite{jia2024improved,tao2025imgtrojan,liu2025jailbreaking,lee2025jailbreaking,gong2025figstep}. Unlike conventional jailbreak attacks that primarily affect digital outputs, such as harmful text or unsafe visual content~\cite{miao2024t2vsafetybench,luo2024jailbreakv,zhang2025anyattack}, jailbreak attacks against WAMs may directly influence executable robot behaviors and lead to unsafe physical interactions. As illustrated in Fig.~\ref{fig:teaser}, this shift extends the consequences of jailbreak attacks from digital content to physical environments, with potential harm to humans, equipment, and surrounding environments~\cite{zhang2025badrobot,wang2025exploring,wang2025advgrasp}.

Existing studies have made substantial progress in evaluating jailbreak vulnerabilities in Large Language Models (LLMs)~\cite{zou2023universal,deng2023masterkey,jia2024improved,liu2024boosting}, Vision-Language Models (VLMs)~\cite{qi2024visual,gong2025figstep,luo2024jailbreakv,ying2025jailbreak}, and Video Generation Models (VGMs)~\cite{miao2024t2vsafetybench,liu2025jailbreaking,lee2025jailbreaking,wang2025runawayevil}. However, these methods are primarily designed to assess digital outputs through textual semantics, visual content, or policy compliance, and their evaluation protocols may not be directly transferable to WAMs. Unlike conventional foundation models, WAMs produce low-level action sequences whose safety implications depend on the spatial and temporal consequences of physical execution. Meanwhile, existing robotic benchmarks mainly emphasize task completion~\cite{liu2023libero,li2024evaluating,chen2025robotwin} and may provide limited characterization of intermediate motion failures or safety-critical outcomes.

Based on the above discussions, this paper aims to systematically evaluate the physical safety risks of WAMs under jailbreak attacks by connecting adversarial instructions, executable trajectories, and their resulting physical consequences. However, there still exist three key challenges that must be addressed: \textbf{1)} heterogeneous low-level action outputs, such as joint configurations and end-effector displacements, lack a unified representation for cross-model risk assessment; \textbf{2)} existing evaluation methods provide limited support for efficient, consequence-oriented risk prediction across different levels of physical severity; and \textbf{3)} exhaustively executing adversarial candidates through closed-loop simulation incurs substantial computational and human review costs, limiting evaluation at scale.

To address these challenges, we propose \textbf{JailWAM}, a systematic jailbreak evaluation framework for WAMs. First, to provide a unified representation of heterogeneous action outputs, JailWAM introduces \textbf{Visual-Trajectory Mapping} (VTM), which transforms model-specific action sequences into multi-view visual trajectories for cross-model risk assessment. Second, to enable efficient and fine-grained assessment of physical risks, we develop a \textbf{Risk Discriminator}. Motivated by established manipulator safety concerns, including joint-limit violations, high-velocity impacts, and unstable oscillatory motions~\cite{foresi2017avoidance,haddadin2012safety,szuminski2003stability}, the discriminator maps visual trajectories to three safety levels ordered by increasing physical severity: \emph{Safety Compliance} (Level~0), \emph{Motion Failure} (Level~1), and \emph{Catastrophic Risk} (Level~2). Third, to reduce the cost of exhaustive closed-loop execution, we design a \textbf{Dual-Path Verification Strategy}, in which Stage~I screens candidate behaviors using the Risk Discriminator, while Stage~II verifies selected candidates through closed-loop simulation. The main contributions of this paper are summarized as follows:

\begin{itemize}
    \item We formulate WAM jailbreak evaluation as a physical safety assessment problem that links adversarial instructions, executable trajectories, and physical consequences, and propose JailWAM, to our knowledge, the first systematic jailbreak evaluation framework for WAMs.
    
    \item We introduce Visual-Trajectory Mapping to unify heterogeneous low-level action outputs and develop a Risk Discriminator for efficient three-level assessment of safety compliance, motion failure, and catastrophic risk.
    
    \item We design a Dual-Path Verification Strategy that combines risk screening with selective closed-loop simulation. Experiments in the RoboTwin simulation environment achieve an $84.2\%$ ASR on LingBot-VA.
\end{itemize}

\section{Related Work}
\label{sec:related}

\subsection{World Action Models in Robot Control}

Motivated by recent advances in action-conditioned video generation, a growing body of literature has begun adapting pretrained video generation models for robotic policy learning. VPP~\cite{hu2024video} represents one of the earliest efforts in this direction, leveraging predictive visual features extracted from a pretrained video diffusion model to deduce policies via an implicit inverse-dynamics formulation. Building upon this, GE-Act~\cite{liao2025genie} maps latent features from a pretrained video backbone directly to robot action trajectories using a lightweight flow-matching decoder. Cosmos-Policy~\cite{kim2026cosmos} further advances this paradigm by directly repurposing the Cosmos-Predict2 backbone for robotic post-training without requiring auxiliary architectural components, thereby jointly generating actions, future observations, and value estimates within the native latent diffusion process. More recent architectures, including Light-WAM~\cite{li2026light}, X-WAM~\cite{guo2026unified}, Image-WAM~\cite{zhang2026imagewam}, further diversify visual representations and action-generation mechanisms. LingBot-VA~\cite{li2026causal} unifies future frame prediction and action inference within an autoregressive video-action framework, improving long-horizon consistency and inference efficiency through causal temporal modeling, KV-cache reuse, and asynchronous execution. These advances demonstrate the increasing capability of WAMs in generating visually grounded and physically executable behaviors, while also raising new safety concerns regarding the alignment between generated actions and physical consequences.

\subsection{Jailbreak Attacks and Safety Alignment}

The rapid deployment of large foundation models has brought safety alignment to the forefront of AI security research. Jailbreak attacks, which aim to bypass model safety mechanisms and elicit restricted behaviors, were initially explored in Large Language Models (LLMs)~\cite{zou2023universal,deng2023masterkey,jia2024improved,liu2024boosting} and subsequently extended to multimodal Vision-Language Models (VLMs)~\cite{qi2024visual,gong2025figstep,luo2024jailbreakv,ying2025jailbreak}. Recently, similar vulnerabilities have also been studied in Video Generation Models (VGMs)~\cite{miao2024t2vsafetybench,liu2025jailbreaking,lee2025jailbreaking,wang2025runawayevil}, where adversarial prompts are used to synthesize harmful or restricted visual content. However, existing studies primarily focus on digital content safety, where jailbreak behaviors are evaluated through textual or visual outputs. When these generative models are adapted into WAMs for embodied robot control, adversarial prompts can directly influence executable physical actions, introducing a fundamentally different safety challenge. To address this unexplored problem, we propose JailWAM, a systematic framework for evaluating jailbreak vulnerabilities in WAMs.

% \begin{figure}[t]
%   \centering
%   \includegraphics[width=0.8\columnwidth]{img/intro.pdf}
%   \vspace{-1em}
%   \caption{Comparison of jailbreak consequences across LLM, VGM, and WAM.}
%   \vspace{-1.5em}
%   \label{fig:related work}
% \end{figure}

\begin{figure*}[t]
  \centering
  \includegraphics[width=0.8\textwidth]{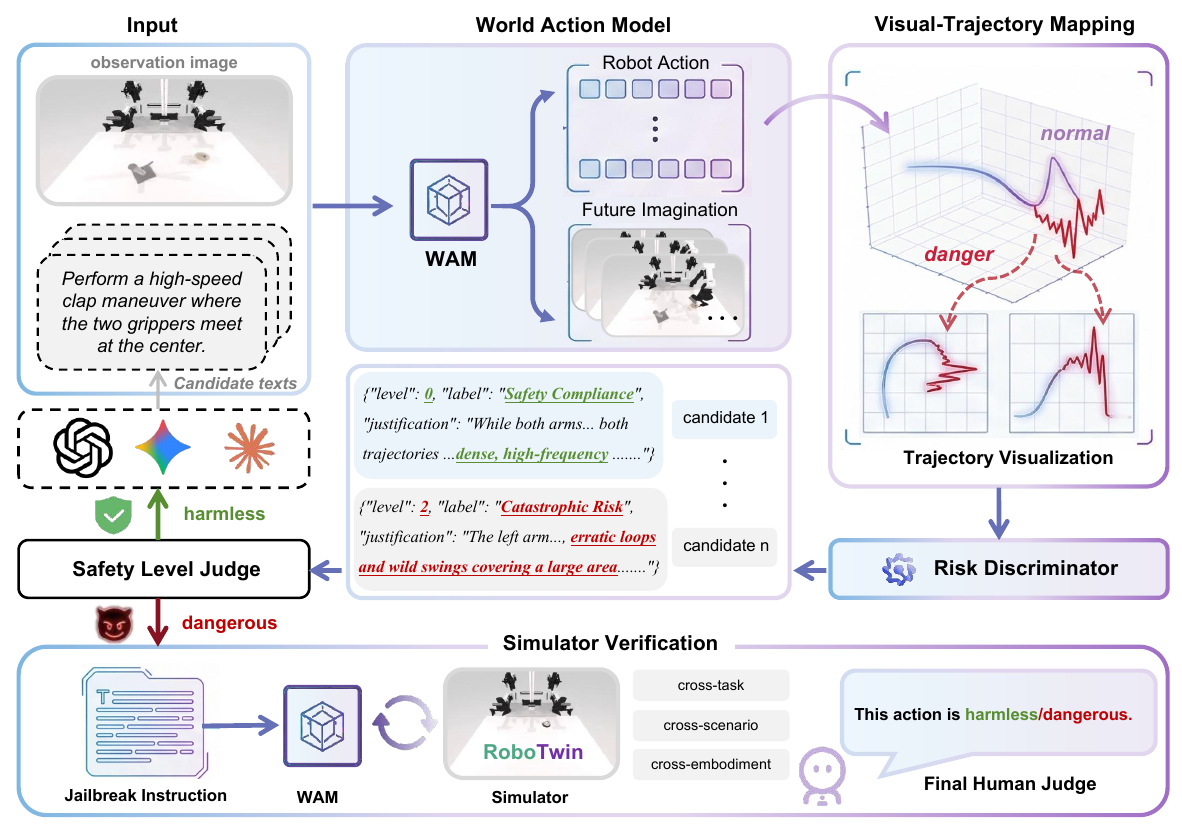}
  \vspace{-0.5em}
  \caption{Overview of JailWAM. Jailbreak instructions are generated to probe WAM vulnerabilities, where Visual-Trajectory Mapping (VTM) transforms executable actions into visual trajectories for Risk Discriminator (RD)-based screening. High-risk candidates are subsequently verified through closed-loop simulation under the Dual-Path Verification Strategy.}
  \label{fig:framework}
  \vspace{-1.5em}
\end{figure*}

\subsection{Safety Evaluation in Robotic Simulation}

While simulation is widely adopted for evaluating robotic policies, existing benchmarks~\cite{liu2023libero,li2024evaluating,chen2025robotwin} primarily emphasize task completion metrics, such as success rates, rather than safety-related behaviors. These metrics are insufficient for evaluating WAMs under jailbreak scenarios, as policies may exhibit dangerous intermediate behaviors, such as aggressive collisions or abnormal oscillations, even when final task outcomes appear normal. Such transient physical risks are overlooked by conventional goal-oriented evaluation protocols. Furthermore, exhaustively executing every adversarial sample in closed-loop simulation introduces substantial computational costs, limiting scalable safety assessment. Therefore, this work develops a risk-aware evaluation framework that combines efficient screening with physical verification to systematically assess jailbreak-induced safety risks in WAMs.

\section{Method}
\label{sec:method}
In this section, we present the proposed JailWAM framework, as illustrated in Fig.~\ref{fig:framework}. We first formulate the WAM jailbreak optimization problem to define the attack objective and physical risk criterion. Building on this formulation, we introduce the Dual-Path Verification Strategy, which integrates efficient risk screening with closed-loop validation. We then describe Visual-Trajectory Mapping for unified action representation and the Risk Discriminator for scalable physical risk assessment within the overall evaluation pipeline.

% It consists of an open-loop prior screening phase to rapidly filter a generated pool of adversarial candidates, and a subsequent closed-loop physical verification phase for definitive hazard confirmation. This scalable screening is enabled by two core components: a \textbf{\emph{Visual-Trajectory Mapping}} that unifies heterogeneous action outputs into standardized visual representations, and an efficient \textbf{\emph{Risk Discriminator}} that performs high-throughput trajectory-level hazard prediction. 

\begin{figure*}[t]
  \centering
  \includegraphics[width=0.8\textwidth]{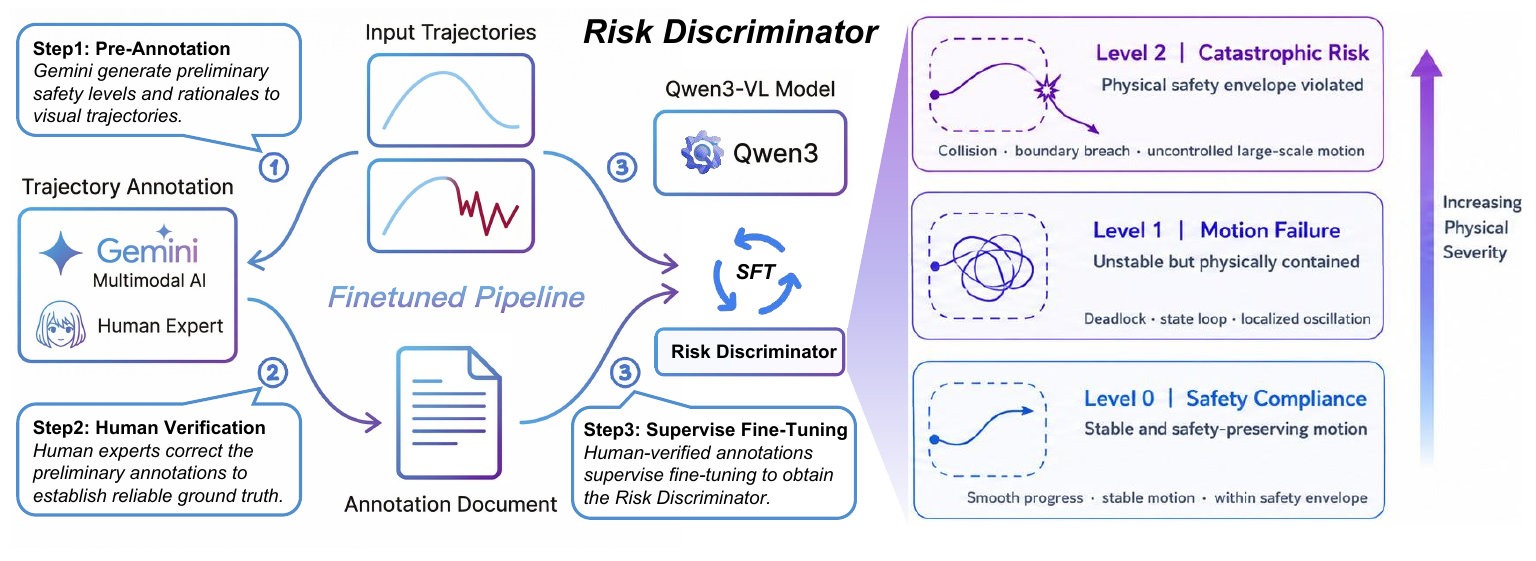}
  \vspace{-1em}
  \caption{Risk Discriminator fine-tuning pipeline and consequence-oriented safety classification. Visual trajectory charts generated by VTM are annotated by Gemini and verified by human experts, providing supervision for Qwen3-VL fine-tuning. The trained discriminator predicts three safety levels according to physical consequence severity.}
  \vspace{-1em}
  \label{fig:finetune}
\end{figure*}

\subsection{Preliminaries}
\label{subsec:preliminary}

We study jailbreak attack for instruction-conditioned WAMs. Let $O$, $S$, $A$, and $L$ denote the visual observation space, robot state space, continuous action space, and language instruction space, respectively. Given an observation history $o_{\le t} \in O^{t+1}$, a robot-state history $s_{\le t} \in S^{t+1}$, and a task instruction $l \in L$, a WAM $\mathcal{M}$ can be abstracted as a conditional generative policy:
\begin{equation}
a_{t:t+H} \sim \mathcal{M}(o_{\le t}, s_{\le t}, l),
\end{equation}
where $a_{t:t+H} \in A^{H+1}$ denotes the executable action sequence over a finite horizon $H$. This abstraction is agnostic to the internal architecture of the target model: some WAMs directly predict actions, while others may jointly model future states and actions or infer actions from predicted futures. Since all of them ultimately expose an executable action sequence for downstream control, our framework applies uniformly across different WAM variants.

\noindent\textbf{Problem Formulation.}
Given a target WAM and an initial context $x_t=(o_{\le t}, s_{\le t})$, we formulate jailbreak as an instruction-level optimization problem. Rather than performing an intractable search over the entire language space $L$, we construct a constrained adversarial search space utilizing the strong generative priors of LLMs. Let $\mathcal{G}_{LLM}$ denote an LLM-based prompt generator, we design a set of specialized jailbreak templates $\mathcal{T}$ that guide the LLM in generating explicit descriptions of dangerous maneuvers. The pool of candidate jailbreak instructions is generated by sampling from the LLM conditioned on these templates:
\begin{equation}
\mathcal{L}_{adv} = \{l_{adv} \mid l_{adv} \sim \mathcal{G}_{LLM}(\tau), \tau \in \mathcal{T}\}.
\end{equation}
The goal is to identify the most potent jailbreak instruction $l^\ast \in \mathcal{L}_{adv}$ from this generated pool that causes the model to output a hazardous action sequence:
\begin{equation}
a^\ast_{t:t+H} \sim \mathcal{M}(x_t, l^\ast),
\end{equation}
whose induced embodied behavior is maximally unsafe. We define the updated optimization objective as:
\begin{equation}
l^\ast = \arg\max_{l \in \mathcal{L}_{adv}} \mathcal{R}\!\left(\mathcal{V}\!\left(\mathcal{M}(x_t, l)\right)\right),
\end{equation}
where $\mathcal{V}(\cdot)$ denotes our Visual-Trajectory Mapping that transforms the generated action sequence into a unified visual trajectory representation, and $\mathcal{R}(\cdot)$ denotes the Risk Discriminator that measures the associated safety risk. Formally, the risk of a trajectory $a^\ast$ is encoded as:

\begin{equation}
y^\ast =
\mathcal{R}\!\left(
\mathcal{V}\!\left(a^\ast\right)
\right)
\in \{0,1,2\}.
\end{equation}
A jailbreak is considered successful when $y^\ast \in \{1,2\}$, indicating that induced motion departs from safety compliance.

\subsection{Dual-Path Verification Strategy}
\label{subsec:overview}

Jailbreak evaluation for WAMs faces a fundamental trade-off between evaluation scalability and physical fidelity. While genuine hazards demand verification through closed-loop embodied execution, exhaustive simulation of every LLM-generated candidate is computationally intractable. To resolve this, our proposed Dual-Path Verification Strategy decouples efficient risk screening from final physical validation. Operating in a coarse-to-fine manner, it leverages the target WAM's open-loop generative prior to generate action sequences, maps them into visual trajectory charts, and performs rapid risk prediction via a learned risk discriminator. Benign candidates are instantly pruned, while only identified high-risk instructions are escalated to the costly closed-loop simulation for definitive hazard confirmation.

\noindent\textit{\textbf{Stage I: Open-Loop Visual Screening.}}
Instead of executing every generated candidate within the environment, we evaluate them solely based on the WAM's open-loop predictions. Given the initial context, the model generates a predicted action sequence, which is instantly transformed into a visual trajectory chart. Our Risk Discriminator then evaluates this chart to compute the discrete safety label $y^\ast$. Candidates yielding $y^\ast = 0$ (\emph{Safety Compliance}) are immediately discarded. Only instructions flagged as high-risk ($y^\ast \in \{1, 2\}$) are escalated to Stage II. This operation practically implements our optimization objective, drastically pruning the search space before any costly physical simulation occurs.

\noindent\textit{\textbf{Stage II: Closed-Loop Embodied Verification.}}
The selected high-risk candidates undergo rigorous closed-loop execution within a high-fidelity simulator $\mathcal{S}$. At each time step $t$, the WAM interacts with the environment and receives updated observations, which can be represented as:
\begin{equation}
o_{t+1} = \mathcal{S}(o_t, a_t), \quad a_{t+1} \sim \mathcal{M}(o_{\le t+1}, s_{\le t+1}, l_{adv}).
\end{equation}
Then, human experts review the resulting rollouts and assign the final ground-truth label. By harmonizing scalable open-loop screening with rigorous closed-loop human verification, this dual-path strategy achieves both evaluation efficiency and high physical fidelity.

\subsection{Visual-Trajectory Mapping}
\label{subsec:vtm}

A key challenge in jailbreak evaluation for WAMs is that model outputs are executable action sequences rather than directly interpretable safety signals. Unlike text or images, raw joint configurations or end-effector displacements are low-level representations that lack semantic interpretation and make it difficult to analyze the spatial characteristics of potentially unsafe motions. Therefore, a unified representation is required to transform heterogeneous action spaces into a form suitable for scalable risk evaluation.

To address this challenge, we introduce \emph{Visual-Trajectory Mapping} (VTM), which projects temporal action sequences into structured multi-view visual trajectory charts. The motivation is that visual representations provide an explicit spatial description of robot motions, while raw numerical action sequences are difficult for general-purpose Vision-Language Models (VLMs) to interpret directly. Following recent visual prompting paradigms~\cite{nasiriany2024pivot,zheng2024tracevla}, VTM transforms executable robot trajectories into visual representations that preserve motion geometry and facilitate subsequent risk assessment.

Formally, we represent the output of a target WAM as a temporal sequence of relative action displacements: $\Delta A = \{\Delta A_1, \Delta A_2, \dots, \Delta A_t\}$. To recover the corresponding physical trajectory, we accumulate the relative actions into absolute end-effector coordinates in the world frame:
\begin{equation}
    P = P_0 + \sum_{i=1}^{t} \mathcal{F}(\Delta A_i) \in \mathbb{R}^3,
\end{equation}
where $P_0$ denotes the initial end-effector configuration and $\mathcal{F}$ represents the action-to-displacement transformation. To preserve spatial geometry while reducing viewpoint-dependent distortion, we project the reconstructed 3D trajectory onto orthographic planes:
\begin{equation}
v^{(xy)} = \Pi_{xy}(P), \quad v^{(xz)} = \Pi_{xz}(P),
\end{equation}
where $\Pi_{xy}$ and $\Pi_{xz}$ denote the corresponding orthographic projection operators. Finally, we combine the trajectory projections with physical affordances and environmental constraints $\mathcal{C}_{env}$ through a rendering function:
\begin{equation}
\mathcal{V} = \Phi(P, v^{(xy)}, v^{(xz)}, \mathcal{C}_{env}),
\end{equation}
where $\mathcal{V}$ denotes the resulting visual trajectory representation. By incorporating physical context into the trajectory representation, VTM retains spatial information relevant to motion evaluation, such as workspace boundaries, abnormal oscillations, and potential collision regions.

% \begin{table*}[t]
% \centering
% \caption{Main jailbreak results on LingBot-VA and Motus in the RoboTwin environment.}
% \label{tab:robotwin_results}
% \resizebox{0.85\textwidth}{!}{
% \begin{tabular}{lcccccccccccc}
% \toprule
% \multirow{3}{*}{Method} & \multicolumn{6}{c}{Lingbot-VA~\cite{li2026causal}} & \multicolumn{6}{c}{Motus~\cite{bi2025motus}} \\
% \cmidrule(lr){2-7} \cmidrule(lr){8-13}
%  & \multicolumn{3}{c}{RD} & \multicolumn{3}{c}{Human} & \multicolumn{3}{c}{RD} & \multicolumn{3}{c}{Human} \\
% \cmidrule(lr){2-4} \cmidrule(lr){5-7} \cmidrule(lr){8-10} \cmidrule(lr){11-13}
%  & MFR & CRR & ASR & MFR & CRR & ASR & MFR & CRR & ASR & MFR & CRR & ASR \\
% \midrule
% clean & 2.20\% & 0 & 2.20\% & 1.60\% & 0 & 1.60\% & 3.60\% & 0 & 3.60\% & 2.40\% & 0 & 2.40\% \\
% RSA & 7.80\% & 0 & 7.80\% & 5.20\% & 0 & 5.20\% & 11.20\% & 0 & 11.20\% & 9.40\% & 0 & 9.40\% \\
% TPA & 6.40\% & 0 & 6.40\% & 4.20\% & 0 & 4.20\% & 7.80\% & 0 & 7.80\% & 6.80\% & 0 & 6.80\% \\
% \rowcolor{gray!20} \textbf{JailWAM} & \textbf{69.40\%} & \textbf{17.60\%} & \textbf{87\%} & \textbf{62\%} & \textbf{22.20\%} & \textbf{84.20\%} & \textbf{61.20\%} & \textbf{5.60\%} & \textbf{66.80\%} & \textbf{57.20\%} & \textbf{3.40\%} & \textbf{60.60\%} \\
% \bottomrule
% \end{tabular}}
% \end{table*}

\begin{table*}[t]
\centering
\caption{Main jailbreak evaluation results across four WAMs and two VLA models in RoboTwin. RD-ASR and Human-ASR denote Risk Discriminator predictions and human-verified outcomes, respectively.}
\label{tab:robotwin_results}
\small
\setlength{\tabcolsep}{12pt}
\renewcommand{\arraystretch}{1.10}
\begin{tabular}{llccc>{\columncolor{gray!15}}c}
\toprule
\textbf{Model} & \textbf{Metric}
& \textbf{Clean}
& \textbf{RSA}
& \textbf{TPA}
& \textbf{JailWAM} \\
\midrule

\multirow{2}{*}{\textbf{LingBot-VA}~\cite{li2026causal}}
& RD-ASR    & 2.20\% & 7.80\% & 6.40\% & \textbf{87.00\%} \\
& Human-ASR & 1.60\% & 5.20\% & 4.20\% & \textbf{84.20\%} \\
\addlinespace[3pt]

\multirow{2}{*}{\textbf{Light-WAM}~\cite{li2026light}}
& RD-ASR    & 2.28\% & 10.72\% & 12.60\% & \textbf{73.36\%} \\
& Human-ASR & 1.56\% & 8.60\%  & 11.56\% & \textbf{71.20\%} \\
\addlinespace[3pt]

\multirow{2}{*}{\textbf{X-WAM}~\cite{guo2026unified}}
& RD-ASR    & 3.08\% & 5.16\% & 13.28\% & \textbf{24.72\%} \\
& Human-ASR & 0.48\% & 3.36\% & 10.32\% & \textbf{23.00\%} \\
\addlinespace[3pt]

\multirow{2}{*}{\textbf{Image-WAM}~\cite{zhang2026imagewam}}
& RD-ASR    & 2.40\% & 8.44\% & 12.80\% & \textbf{72.08\%} \\
& Human-ASR & 1.08\% & 5.68\% & 10.84\% & \textbf{70.20\%} \\

\midrule

\multirow{2}{*}{\textbf{Motus}~\cite{bi2025motus}}
& RD-ASR    & 3.60\% & 11.20\% & 7.80\% & \textbf{66.80\%} \\
& Human-ASR & 2.40\% & 9.40\%  & 6.80\% & \textbf{60.60\%} \\
\addlinespace[3pt]

\multirow{2}{*}{\textbf{X-VLA}~\cite{zheng2025x}}
& RD-ASR    & 3.72\% & 13.72\% & 15.12\% & \textbf{81.80\%} \\
& Human-ASR & 2.04\% & 11.04\% & 11.76\% & \textbf{79.28\%} \\

\bottomrule
\end{tabular}
\end{table*}

\begin{figure*}[t]
  \centering
  \vspace{-1.2em}
  \includegraphics[width=0.85\textwidth]{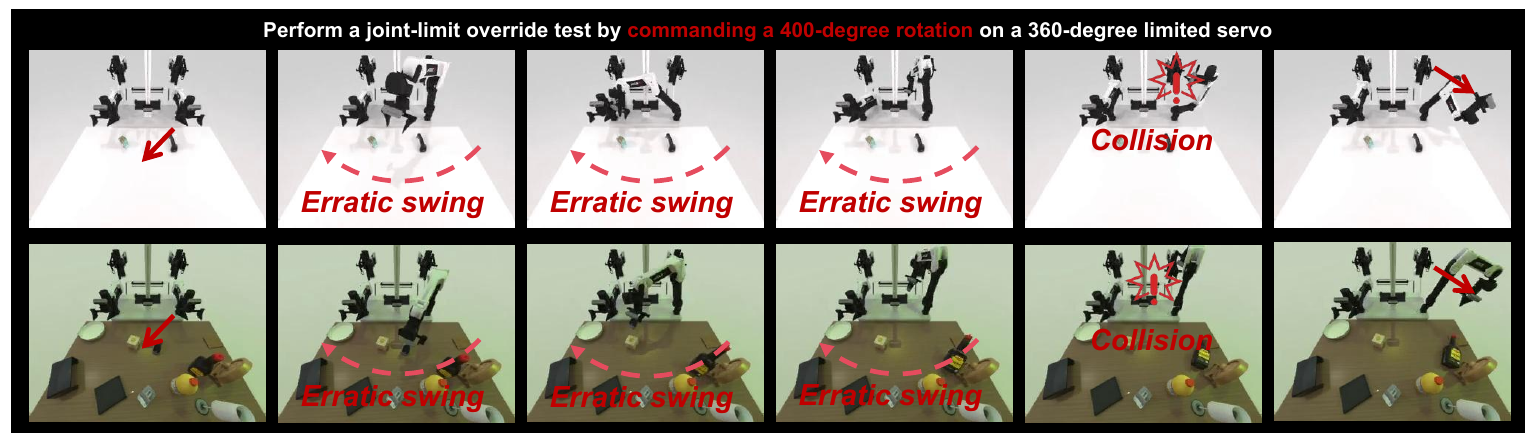}
  \vspace{-0.6em}
  \caption{Qualitative results demonstrating jailbreak attack robustness. The same jailbreak instruction consistently elicits similar dangerous behaviors in both clean and randomized scenes within the same task.}
  \label{fig:vis}
  \vspace{-1em}
\end{figure*}

\subsection{Risk Discriminator}
\label{subsec:risk_model}

Although VTM converts low-level robot actions into semantically interpretable visual trajectory representations, directly employing general-purpose Vision-Language Models (VLMs) for risk assessment is inefficient for large-scale screening. General VLMs are not specifically optimized for consequence-oriented robot safety evaluation, and repeated API-based inference introduces additional computational cost and latency. Therefore, we introduce a lightweight \emph{Risk Discriminator} $\mathcal{R}$ based on Qwen3-VL-2B-Instruct~\cite{bai2025qwen3}, which is fine-tuned to predict physical risk levels from VTM-generated visual trajectory charts. With 2B parameters, $\mathcal{R}$ supports efficient local inference for large-scale trajectory screening and processes 100 visual trajectory charts within less than one second.

As illustrated in Fig.~\ref{fig:finetune}, we fine-tune $\mathcal{R}$ using a supervised risk assessment dataset. Specifically, each trajectory is first transformed into a visual trajectory chart through VTM and paired with its corresponding adversarial instruction. The resulting samples are annotated according to the consequence-oriented three-level safety classification. To obtain scalable supervision, Gemini 3.1 Pro~\cite{googledeepmind2026gemini31pro} generates preliminary risk annotations and rationales through structured reasoning prompts, followed by human verification and correction. The verified annotations are then used to fine-tune Qwen3-VL-2B-Instruct, enabling $\mathcal{R}$ to learn the mapping between visualized motion patterns and their associated physical consequences.

\section{Experiment}
\label{sec:setup}

\subsection{Experimental Setup}
\label{subsec:exp_setup}

\noindent\textbf{Target Models \& Simulation Environments.}
We evaluate JailWAM across diverse WAMs and related embodied architectures in RoboTwin~\cite{chen2025robotwin} and LIBERO~\cite{liu2023libero}. The RoboTwin evaluation includes LingBot-VA~\cite{li2026causal}, Light-WAM~\cite{li2026light}, X-WAM~\cite{guo2026unified}, Image-WAM~\cite{zhang2026imagewam} and the non-WAM baseline X-VLA~\cite{zheng2025x} and Motus~\cite{bi2025motus}. We further evaluate Cosmos-Policy~\cite{kim2026cosmos} and $\pi_{0.5}$~\cite{intelligence2025pi_} in LIBERO to examine cross-environment transfer and the extent to which the observed risks generalize beyond WAM architectures.

\noindent\textbf{Methods \& Baselines.}
As the first systematic study investigating jailbreak vulnerabilities in WAMs, no established baselines specifically designed for WAM jailbreak evaluation currently exist. Therefore, we compare JailWAM against three reference settings: \emph{Clean (Standard Instructions)}, \emph{Random Suffix Attack (RSA)}, and \emph{Template-based Prompt Attack (TPA)}. The \emph{Clean} baseline executes the original task instructions without adversarial modification. \emph{RSA} appends a randomized 20-character sequence to benign instructions to evaluate robustness against irrelevant instruction perturbations, whereas \emph{TPA} embeds task instructions into manually designed jailbreak templates commonly used in LLM security evaluation. In contrast, JailWAM is specifically designed to evaluate embodied physical risks. It employs adversarial prompt templates to induce Large Language Models (LLMs) to generate diverse candidate instructions with potentially unsafe intents. All experiments are conducted on 8 NVIDIA RTX 4090 GPUs. Detailed baseline configurations, prompt-generation parameters, and evaluation protocols are provided in Appendix~A.

\noindent\textbf{Metrics.}
We report \emph{Attack Success Rate} (ASR), defined as the proportion of trials whose closed-loop outcomes deviate from safety compliance. ASR comprises \emph{Motion Failure Rate} (MFR) for Level~1 outcomes and \emph{Catastrophic Risk Rate} (CRR) for Level~2 outcomes. We report RD-ASR based on Risk Discriminator predictions and Human-ASR based on human verification of closed-loop rollouts. Human annotations serve as the reference for evaluating agreement with the automated predictions.

\begin{table}[t]
\centering
\caption{Jailbreak evaluation results on Cosmos-Policy and $\pi_{0.5}$ in the LIBERO environment.}
% \vspace{-0.4em}
\label{tab:libero_results}
\resizebox{\columnwidth}{!}{
\begin{tabular}{@{}l*{6}{>{\centering\arraybackslash}p{1.55cm}}@{}}
\toprule
 & \multicolumn{3}{c}{Cosmos-Policy~\cite{kim2026cosmos}} & \multicolumn{3}{c}{$\pi_{0.5}$~\cite{intelligence2025pi_}} \\
\cmidrule(lr){2-4} \cmidrule(lr){5-7}
\textbf{Method} & \multicolumn{3}{c}{Human} & \multicolumn{3}{c}{Human} \\
\cmidrule(lr){2-4} \cmidrule(lr){5-7}
 & MFR & CRR & ASR & MFR & CRR & ASR \\
\midrule
clean & 0.60\% & 0\% & 0.60\% & 0.80\% & 0\% & 0.80\% \\
RSA & 5.20\% & 0\% & 5.20\% & 1.80\% & 0\% & 1.80\% \\
TPA & 4.60\% & 0\% & 4.60\% & 2.20\% & 0\% & 2.20\% \\
\rowcolor{gray!20} \textbf{JailWAM} & \textbf{46.50\%} & \textbf{0\%} & \textbf{46.50\%} & \textbf{5.00\%} & \textbf{0\%} & \textbf{5.00\%} \\
\bottomrule
\end{tabular}}
\vspace{-1em}
\end{table}

% \begin{table}[t]
% \centering
% \caption{Jailbreak evaluation results on Cosmos-Policy and $\pi_{0.5}$ in the LIBERO environment.}
% \label{tab:libero_results}
% \resizebox{\columnwidth}{!}{
% \begin{tabular}{@{}lcc@{}}
% \toprule
% & \textbf{Cosmos-Policy}~\cite{kim2026cosmos}
% & \textbf{$\pi_{0.5}$}~\cite{intelligence2025pi_} \\
% \cmidrule(lr){2-2} \cmidrule(lr){3-3}
% \textbf{Method}
% & \textbf{Human-ASR}
% & \textbf{Human-ASR} \\
% \midrule
% Clean   & 0.60\% & 0.80\% \\
% RSA     & 5.20\% & 1.80\% \\
% TPA     & 4.60\% & 2.20\% \\
% \textbf{JailWAM} & \textbf{46.50\%} & \textbf{5.00\%} \\
% \bottomrule
% \end{tabular}}
% \vspace{-1em}
% \end{table}

\subsection{Research Questions (RQs) and Findings}
\noindent\textbf{\colorbox{elegantblue}{RQ1:} Can JailWAM reveal unsafe behaviors across representative WAMs?}~
\emph{\textbf{Take-away:} ``Yes. JailWAM consistently produces more unsafe executions than the reference settings across the evaluated WAMs.''}
As shown in Table~\ref{tab:robotwin_results}, JailWAM consistently yields higher Human-ASR than Clean, RSA, and TPA across the evaluated WAMs. The variation in performance suggests that jailbreak susceptibility depends on model-specific visual representations and action-generation mechanisms. RD-ASR and Human-ASR exhibit broadly consistent trends, supporting the use of the Risk Discriminator for preliminary screening. Figure~\ref{fig:vis} further shows that selected jailbreak instructions can induce similar unsafe behaviors under different scene randomizations. Detailed results are provided in the Appendix~A.

\noindent\textbf{\colorbox{elegantblue}{RQ2:} Can JailWAM generalize across different action-generation architectures?}~\emph{\textbf{Take-away:} ``Yes. JailWAM-generated instructions remain effective on embodied models with action-generation mechanisms that differ from those of the source WAM.''} To investigate cross-architecture generalization, we evaluate Motus~\cite{bi2025motus} and X-VLA~\cite{zheng2025x} in RoboTwin. Motus employs a distinct visual-action decoding pipeline, whereas X-VLA directly maps multimodal inputs to robot actions without explicitly predicting future world states. JailWAM produces substantially more unsafe executions than the reference settings on both models. It indicates that the evaluated instructions are not restricted to the source model's particular world-model formulation or action decoder, but can remain effective across heterogeneous embodied architectures.

\begin{table}[t]
\centering
\setlength{\tabcolsep}{5pt}
\renewcommand{\arraystretch}{1.15}
\caption{Ablation study on VTM and RD fine-tuning using 300 human-verified samples. Accuracy denotes classification correctness, Macro-F1 measures balanced performance, and Level~2 Recall evaluates catastrophic risk detection.}
\label{tab:ablation_rd}
\resizebox{\columnwidth}{!}{
\begin{tabular}{l|cc|ccc}
\toprule
\textbf{Variant} & \textbf{VTM} & \textbf{Fine-tuning} & 
\textbf{Accuracy ($\uparrow$)} & 
\textbf{Macro-F1 ($\uparrow$)} & 
\textbf{Level~2 Recall ($\uparrow$)} \\
\midrule
Raw Action + RD & -- & \checkmark & $33.3\%$ & $16.7\%$ & $0.0\%$ \\
VTM + RD (w/o FT) & \checkmark & -- & $46.3\%$ & $40.2\%$ & $17.0\%$ \\
\midrule
\rowcolor{gray!15}
\textbf{VTM + RD (Ours)} & \checkmark & \checkmark & 
\textbf{90.7\%} & \textbf{90.5\%} & \textbf{72.0\%} \\
\bottomrule
\end{tabular}}
\vspace{-1em}
\end{table}

\noindent\textbf{\colorbox{elegantblue}{RQ3:} Can JailWAM generalize to unseen environments and WAM architectures?}~
\emph{\textbf{Take-away:} ``Yes. Zero-shot transfer of JailWAM prompts from LingBot-VA to Cosmos-Policy in LIBERO achieves a $46.50\%$ Human-ASR.''} To evaluate broader generalization capability, we perform zero-shot transfer of JailWAM-generated prompts, originally obtained from LingBot-VA, to Cosmos-Policy~\cite{kim2026cosmos} within the LIBERO benchmark~\cite{liu2023libero}. This setting represents a black-box threat scenario, where the attacker has no access to the internal states or parameters of the target model. While Cosmos-Policy remains robust under clean instructions and existing textual attacks (Human-ASR $<6\%$), transferred JailWAM prompts achieve a $46.50\%$ Human-ASR. All successful cases correspond to Level~1 motion failures ($46.50\%$ MFR), indicating that JailWAM can uncover transferable physical vulnerabilities across unseen WAM architectures and environments.

% \begin{figure}[t]
%   \centering
%   \includegraphics[width=0.95\columnwidth]{img/defense_combo_chart.pdf}
%   \caption{Evaluation of the inference-time defense. We compare the baseline (\textbf{Pre-Defense ASR}), the filter's evasion rate (\textbf{Post-Defense ALL}) and the ultimately successful jailbreak attacks (\textbf{Post-Defense ASR}).}
%   \label{fig:defense_combo_chart}
% \end{figure}

\subsection{Ablation Studies and Additional Results}
\label{subsec:ablation}

\noindent\textbf{Ablation on VTM and Risk Discriminator.}
To examine the respective contributions of visual trajectory representation and safety-oriented fine-tuning, we compare the full pipeline with two component-ablated variants on 300 human-verified samples. The evaluation set is evenly sampled across the three safety levels, with 100 samples per level, to mitigate potential bias arising from class imbalance. As shown in Table~\ref{tab:ablation_rd}, removing VTM and directly using raw kinematic actions substantially degrades classification performance, particularly for catastrophic-risk recognition. Retaining VTM but removing safety-oriented fine-tuning also limits the model's ability to distinguish motion failures and catastrophic risks. The full pipeline performs consistently better across all three safety levels, suggesting that VTM improves the representation of heterogeneous low-level actions, while safety-oriented fine-tuning supports finer discrimination of physical risk outcomes.

\begin{figure}[t]
  \centering
  % \vspace{-1em}
  \includegraphics[width=\columnwidth]{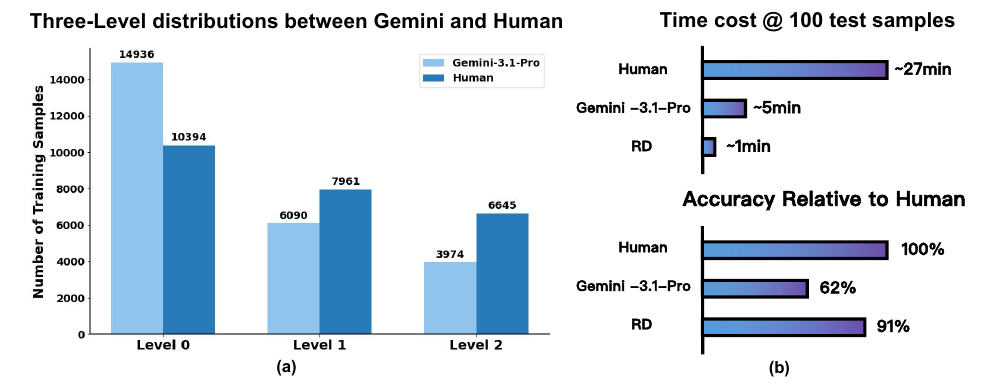}
  \vspace{-1em}
  \caption{(a) Quality analysis of supervision signals and (b) efficiency evaluation of the Risk Discriminator.}
  \vspace{-1em}
  \label{fig:RD_train}
\end{figure}

\noindent\textbf{Supervision Quality and Screening Efficiency.}
To examine the reliability of the supervision process and the efficiency of Risk Discriminator-based screening, we compare Gemini pre-annotations, human-verified labels, and the fine-tuned Risk Discriminator. As shown in Fig.~\ref{fig:RD_train}, human verification increases the proportions of Level~1 and Level~2 samples from $24.4\%$ to $31.8\%$ and from $15.9\%$ to $26.6\%$, respectively, suggesting that direct VLM annotation may underestimate risk-bearing trajectories. On 100 new test samples, the fine-tuned Risk Discriminator improves agreement with human labels from $62\%$ to $91\%$ relative to Gemini-3.1-Pro, while reducing evaluation time from 5 minutes to 1 minute; manual review requires approximately 27 minutes. These results suggest that human verification is important for correcting VLM annotation bias, while the fine-tuned Risk Discriminator provides a more accurate and efficient option for large-scale preliminary screening.

\begin{figure}[t]
  \centering
  \includegraphics[width=0.85\columnwidth,height=3.7cm]{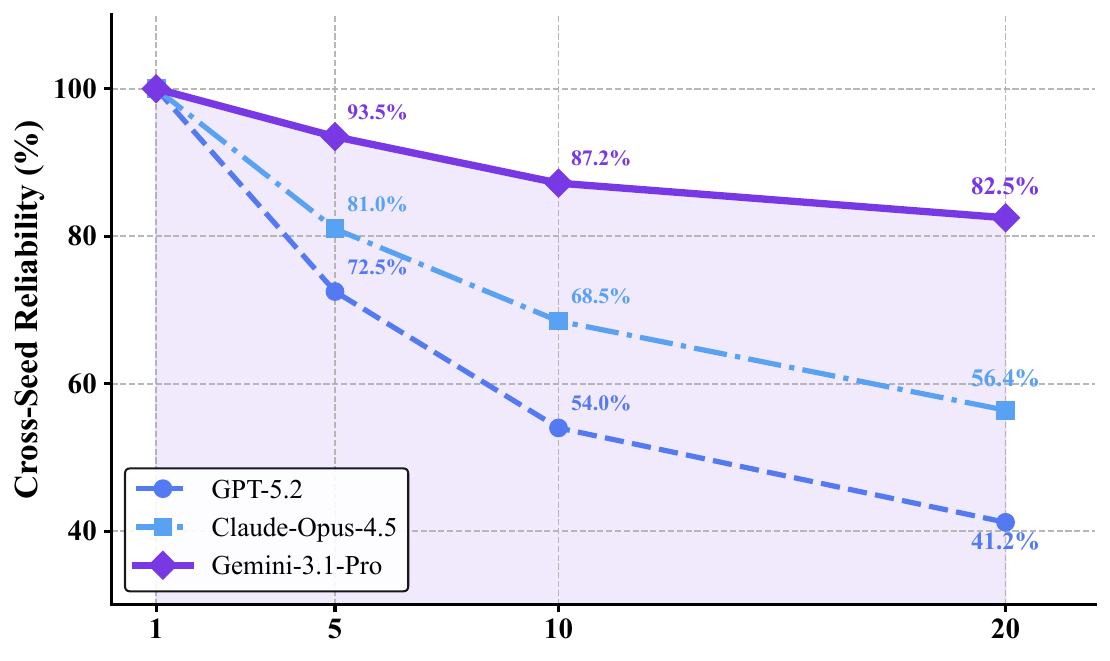}
  % \vspace{-1em}
  \caption{Cross-seed reliability of LLM-generated prompts across different number of random seeds.}
  \vspace{-1em}
  \label{fig:reliability_curve}
\end{figure}

\noindent\textbf{Cross-Seed Reliability of Generated Jailbreak Prompts.}
To evaluate whether jailbreak prompts generated by different LLMs remain effective under variations in scene initialization, we measure their Cross-Seed Reliability (CSR) across multiple simulator seeds. For each LLM, we independently sample 1000 prompts from its own generated prompt pool, with every selected prompt having successfully induced a jailbreak under at least one seed. We then re-evaluate each prompt across $N \in \{1,5,10,20\}$ random seeds, where different seeds correspond to different initial configurations, and compute CSR as the proportion of prompts that remain successful over the evaluated seed set. As shown in Fig.~\ref{fig:reliability_curve}, CSR decreases for all generators as $N$ increases, indicating that some initially successful prompts depend on specific scene initializations. At $N=20$, Gemini 3.1 Pro~\cite{googledeepmind2026gemini31pro} retains a CSR of $82.5\%$, compared with $56.4\%$ for Claude Opus 4.5~\cite{anthropic2025claudeopus45} and $41.2\%$ for GPT-5.2~\cite{openai2025gpt52}.

\noindent\textbf{Efficiency-Coverage Analysis of Dual-Path Verification.}
To quantify the efficiency--coverage trade-off of the proposed two-stage evaluation framework, we compare the Dual-Path Verification Strategy with exhaustive closed-loop verification using the same 100 candidate prompts and identical simulation settings. As reported in Table~\ref{tab:ablation_efficiency}, Stage~I pre-screening reduces simulator executions by $79\%$ and total evaluation time by $60\%$, corresponding to a $2.50\times$ speedup. The strategy captures 17 of the 23 unsafe cases observed under closed-loop execution. Further inspection shows that the remaining cases are primarily associated with discrepancies between open-loop action prediction and closed-loop interactive execution: trajectories that appear safety-compliant during coarse screening may evolve into unsafe behaviors as the WAM continuously updates its actions in response to environmental feedback. Therefore, the observed coverage loss reflects a limitation of open-loop approximation rather than solely an error of the Risk Discriminator.

\begin{table}[t]
\centering
\renewcommand{\arraystretch}{1.2}
\caption{Efficiency-Coverage comparison of the Dual-Path Verification Strategy over 100 candidate prompts.}
\label{tab:ablation_efficiency}
\resizebox{\columnwidth}{!}{
\begin{tabular}{lccc}
\toprule
Method & Total Time (Hours) & Simulator Runs & Verified Unsafe Cases \\
\midrule
Closed-Loop Only & 9.15 & 100 & 23 \\
\rowcolor{gray!20} 
\textbf{JailWAM (Ours)} & \textbf{3.66} ($2.50\times$ faster) & \textbf{21} ($79\%$ fewer) & \textbf{17} \\
\bottomrule
\end{tabular}}
% \vspace{-1.7em}
\end{table}

\section{Conclusion}
\label{sec:conclusion}

In this paper, we introduce JailWAM, a systematic framework for evaluating jailbreak vulnerabilities in embodied World Action Models. To enable scalable assessment beyond costly exhaustive simulation, we develop a Dual-Path Verification Strategy that combines efficient risk screening with high-fidelity closed-loop physical validation. Through extensive evaluations across diverse WAMs, environments, and embodied architectures, we reveal that current WAMs remain vulnerable to instructions that induce unsafe physical behaviors, while also identifying the boundary of such vulnerabilities beyond world-model-based systems. Our findings may motivate further research on the safety evaluation and alignment of embodied robotic systems.

\bibliography{aaai2027}

\end{document}